\documentclass[lettersize,journal]{IEEEtran}
\usepackage{amsmath,amsfonts}
\usepackage{algorithmic}
\usepackage{algorithm}
\usepackage{array}
\usepackage[caption=false,font=footnotesize,labelfont=rm,textfont=rm]{subfig}

\usepackage{textcomp}
\usepackage{stfloats}
\usepackage{url}
\usepackage{verbatim}
\usepackage{graphicx}
\usepackage{cite}

\usepackage[hidelinks]{hyperref}
\usepackage{bbm}
\usepackage{multirow}

\begin{document}

\title{Bridging Domain Gap of Point Cloud Representations via Self-Supervised Geometric Augmentation}

\author{Li Yu, Hongchao Zhong, Longkun Zou, Ke Chen, Pan Gao,~\IEEEmembership{Member,~IEEE}
        % <-this % stops a space
\thanks{This work was supported in part by the National Natural Science Foundation of China under Grant 62002172; and in part by The Startup Foundation for Introducing Talent of NUIST under Grant 2023r131.}% <-this % stops a space
\thanks{Li Yu is with School of Computer Science, Nanjing University of Information Science \& Technology, Nanjing 210044, China, and also with Jiangsu Collaborative Innovation Center of Atmospheric Environment and Equipment Technology (CICAEET), Nanjing University of Information Science \& Technology, Nanjing, China (e-mail: li.yu@nuist.edu.cn).}
\thanks{Hongchao Zhong is with School of Computer Science, Nanjing University of Information Science \& Technology, Nanjing 210044, China (e-mail: 202212200013@nuist.edu.cn).}
\thanks{L. Zou is with the School of Electronic and Information Engineering, South China University of Technology, Guangzhou, 510641, China (e-mail: eelongkunzou@mail.scut.edu.cn).}
\thanks{K. Chen is with the Peng Cheng Laboratory, Shenzhen, China (e-mail: chenk02@pcl.ac.cn).}
\thanks{Pan Gao is with the College of Computer Science and Technology, Nanjing University of Aeronautics and Astronautics, Nanjing 211106, China (e-mail: Pan.Gao@nuaa.edu.cn).}

}

\markboth{Journal of \LaTeX\ Class Files,~Vol.~14, No.~8, August~2021}%
{Shell \MakeLowercase{\textit{et al.}}: A Sample Article Using IEEEtran.cls for IEEE Journals}


\maketitle

\begin{abstract}
Recent progress of semantic point clouds analysis is largely driven by synthetic data (e.g., the ModelNet and the ShapeNet), which are typically complete, well-aligned and noisy-free.
Therefore,  representations of those ideal synthetic point clouds have limited variations in the geometric perspective and can gain good performance on a number of 3D vision tasks such as point cloud classification.
In the context of unsupervised domain adaptation (UDA), representation learning designed for synthetic point clouds can hardly capture domain invariant geometric patterns from incomplete and noisy point clouds.
To address such a problem, we introduce a novel scheme for induced geometric invariance of point cloud representations across domains, via regularizing representation learning with two self-supervised geometric augmentation tasks.
On one hand, a novel pretext task of predicting translation distances of augmented samples is proposed to alleviate centroid shift of point clouds due to occlusion and noises.
On the other hand, we pioneer an integration of the relational self-supervised learning on geometrically-augmented point clouds in a cascade manner, utilizing the intrinsic relationship of augmented variants and other samples as extra constraints of cross-domain geometric features.
Experiments on the PointDA-10 dataset demonstrate the effectiveness of the proposed method, achieving the state-of-the-art performance.
\end{abstract}


\section{Introduction}

\IEEEPARstart{A} point cloud is popularly used to describe object shape with a set of 3D points owing to its simple structure, which encourages a number of 3D vision tasks such as point cloud classification~\cite{pointnet}, \cite{pointnet++}, \cite{dynamic}, 3D detection \cite{detect}.
Recent progress of semantic analysis on point sets is largely driven by synthetic point clouds generated from CAD models (e.g. those in the ModelNet \cite{modelnet} and the ShapeNet \cite{shapenet}), which are typically complete, well-aligned and noise-free.
Geometric variations of ideal synthetic point clouds can significantly be reduced in comparison with those from real-world scenarios, which can be partially occluded and arbitrarily posed. 
In detail, significant differences of geometries in point clouds can be caused by scale variations of objects, self or inter-object occlusion under a single viewpoint, and systematic sensor noises during data acquisition \cite{chen2022quasi}.
%As a result, deep representations of point clouds can hardly be consistent, especially between synthetic and real data.

\begin{figure}[t]
\centering
\includegraphics[height=2.4in,width=3.5in]{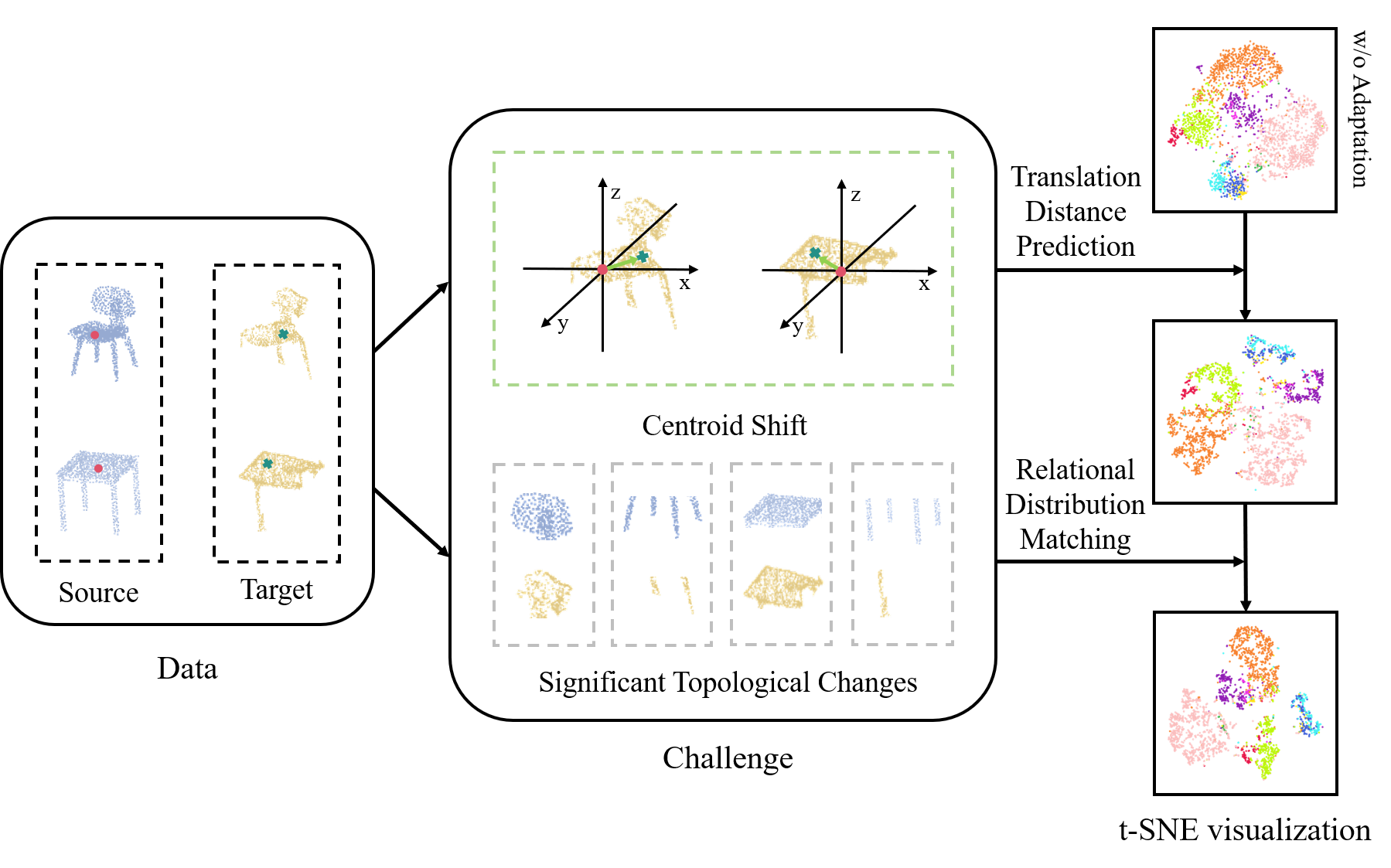}
\caption{
The illustration of resulting t-SNE representation space with and w/o our proposed method for point cloud domain adaptation. The proposed method not only employ translation distance prediction to alleviate centroid shift of point clouds due to occlusion and noises, but also utilize relational learning to further understand the significant topological changes between source and target domains. 
%The effectiveness of self-supervised geometric augmentation for point cloud domain adaptation. We not only employ translation distance prediction to alleviate centroid shift of point clouds due to occlusion and noises, but also utilize relational learning to further understand the relationships between different geometrically-augmented samples.
} 
\label{ill}
\end{figure}

In the context of unsupervised domain adaptation (UDA) of point cloud classification~\cite{pan2009survey}, the goal of representation learning is to extract domain-invariant geometric patterns from one labeled source domain and another unlabeled target domain, which is supervised by target codes of semantic classes.
Evidently, the aim of the semantic task cannot ensure inducing geometric invariance across domains into point cloud representations \cite{tang2020improving}, which encourages a number of explorations to incorporate geometric information through adversarial training~\cite{liang2021boosting}, \cite{kang2019contrastive}, \cite{dann}, \cite{saito2018maximum}, self-training~\cite{ijcai2023p193}, \cite{zou2018unsupervised}, \cite{zou2021geometry} and self-supervised learning such as rotation prediction~\cite{poursaeed2020self}, scaling factors~\cite{GLRV}, distorted part localization \cite{DefRec} and generation of masked parts \cite{MLSP}. 
Existing pretext tasks concern on either achieving representations' generalization on  rotation and scale changes of objects or incorporating cross-domain local geometric information into representations, but very few work has considered to improve representations by coping with geometric variations of partially-observed point clouds from real scenarios.
\IEEEpubidadjcol

We observe that incomplete and noisy point clouds can lead to \textit{centroid shift} and \textit{changes of the topological structure of objects}, and thus make point cloud representations inconsistent between domains, especially between synthetic and real data, as shown in Figure \ref{ill}.
In this paper, we propose a novel self-supervised regularization scheme of representation learning in the problem of UDA, which can discover domain invariant geometric patterns by predicting centroid shift and consistent relation of augmented point clouds from one instance and other instances.  
On the one hand, in order to address the challenge of centroid shift, this paper for the first time designs a self-supervised translation distance prediction task, predicting the translation distance of the augmented point clouds shifting along the coordinate axes, which thus can improve representation generalization on misaligned point clouds. 
On the other hand, inspired by the ReSSL~\cite{ressl}, we adapt the relational self-supervised learning to the UDA on point cloud classification, but novelly in a cascade manner.
Specifically, our relational self-supervised learning method not only minimizes the relationship distribution of weakly augmented and strongly augmented variants of one sample as~\cite{ressl} to regularize representation learning, but also takes the original sample into consideration with the weakly augmented point clouds to form another pair as an extra relation constraint. 
This strategy effectively extends the decision boundary and promotes the distribution of class centers to be more uniform in feature space, and thus can improve robustness against geometric topology variations and discriminant ability of point cloud representations.
Our scheme follows the GAST~\cite{zou2021geometry} to combine the proposed self-supervised regularization terms with the self-paced self-training. 
We conduct experiments on the widely-used benchmarking PointDA-10 dataset on the problem of 3D UDA, whose results confirm the effectiveness of our proposed method and achieve the state-of-the-art performance.

The contributions of this paper are summarized as follows: 
\begin{itemize} 
    \item We propose a novel scheme to regularize representation learning in the context of 3D UDA on point sets, which can effectively narrow domain gaps via self-supervised geometric augmentation.        
    \item Technically, we design two self-supervised learning tasks, one for translation distance prediction to alleviate centroid shift and another for exploration of the relationship between different instances.
%    We employ relational learning as a self-supervised task for the first time in unsupervised domain adaptation of point clouds, exploring the relationships between different samples.
%    \item We propose a self-supervised task for translation distance prediction, alleviating the issue of centroid shift in point clouds, while also enhancing the model's understanding of individual samples.
    \item Experimental results on the widely-recognized benchmark can demonstrate that our method become the new state-of-the-art for the unsupervised domain adaptation of point cloud classification.
\end{itemize}
Source codes and pre-trained models will be released\footnote{\url{Link-of-source-codes-and-models-to-be-downloaded.}}.

\section{Related Work}
\subsection{Deep Classification on Point Sets}
Point cloud is a set of points, which can represent three-dimensional spatial information simply and directly. And classification of point clouds represents a crucial task in the study of point cloud analysis. However, due to its irregularity and permutation invariance, typical 2D image deep learning methods cannot be directly applied to point clouds. To solve this problem, multiple deep neural networks applied to point clouds have been proposed. PointNet~\cite{pointnet} is the first deep neural network that directly processes the raw point cloud, but it lacks the extraction of local features. PointNet++~\cite{pointnet++} combines global and local geometric information in a hierarchical manner based on PointNet. DGCNN~\cite{dynamic} builds a feature space graph and dynamically updates it to aggregate features. Recently, PointTransformer~\cite{PointTransformer} has implemented the Transformer architecture for point cloud processing, resulting in state-of-the-art performance across multiple benchmarks. LCPFormer~\cite{LCPFormer} proposes a solution that leverages the natural structure of point clouds for message passing between local regions, enhancing their representational comprehensiveness and discriminability.

\subsection{Unsupervised Domain Adaptation}
Unsupervised domain adaptation (UDA) for 2D images has been studied for many years and primarily falls into two categories: minimizing the domain discrepancy proxy~\cite{liang2021boosting,pan2019transferrable,kang2019contrastive,deng2019cluster} and adversarial training~\cite{dann,saito2018maximum,li2021triple,DAFL}, \cite{DSEM}. The former measures the discrepancy through distribution statistics, while the latter aligns feature distributions by playing minimax games at the domain or class level. Additionally, the pseudo-labeling technique~\cite{lee2013pseudo}, \cite{gu2020spherical}, \cite{chen2020self}, \cite{shin2020two} refines the model by generating and utilizing pseudo-labels for target domain data, further reducing the domain gap. Inspired by the work in the image domain, UDA has also been applied in the point cloud field. For instance, PointDAN~\cite{pointdan} employs adversarial training to align features across different domains. ALSDA~\cite{ALSDA} presents an automatic loss function search method to tackle domain discriminator degeneration and cross-domain semantic mismatches in adversarial domain adaptation. DefRec~\cite{DefRec} adopts self-supervised learning to capture informative representation with rich local geometric details. GAST~\cite{zou2021geometry} uses a self-training strategy to further reduce the domain gap, enhancing the accuracy of pseudo-labels through self-paced learning. GLRV~\cite{GLRV} proposes a voting-based pseudo-label generation method, effectively improving the reliability of pseudo-labels. COT~\cite{COT} employs multimodal contrastive learning to better separate different categories and utilizes optimal transport to reduce the domain gap. 

\begin{figure*}[htbp]
\centering
\includegraphics[height=4.0in,width=5.9in]{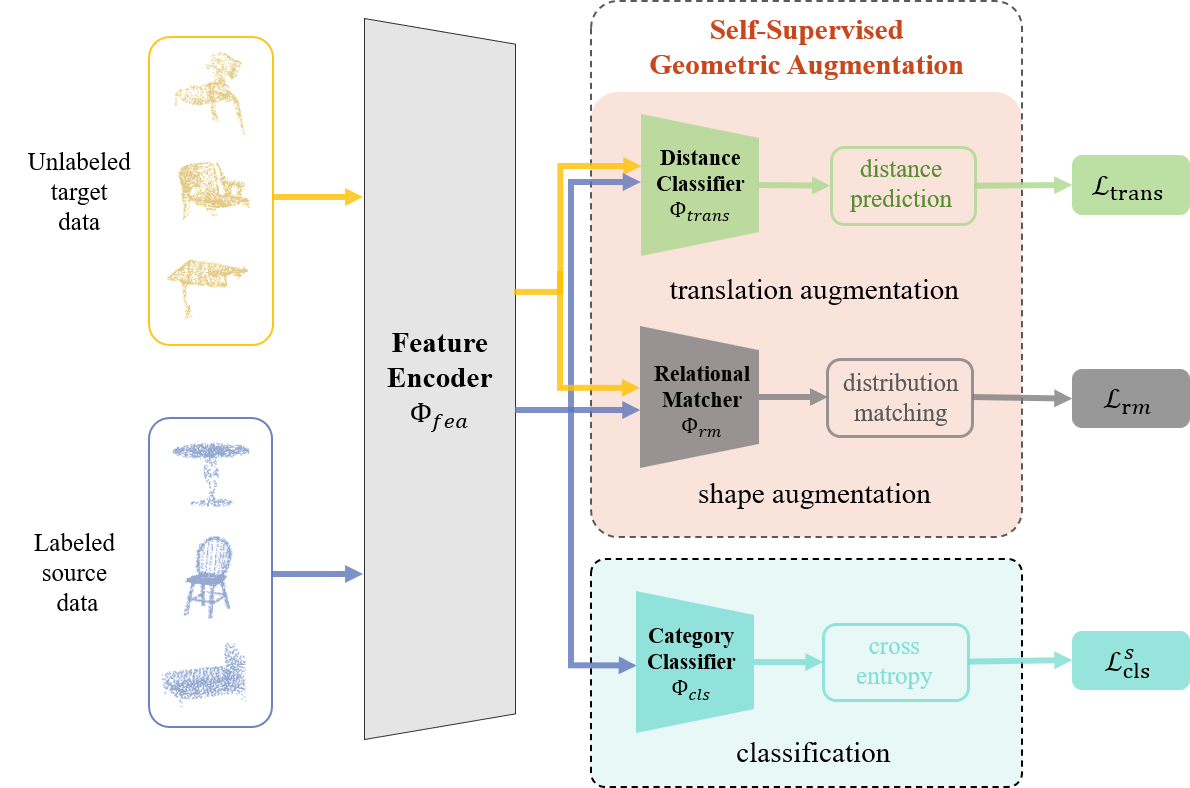}
\caption{The framework of our proposed method for unsupervised domain adaptation on point clouds. The framework comprises three critical components: translation distance prediction to alleviate centroid shift of point clouds, relational modeling to capture relationships between cross-domain samples, and representation learning through supervised learning to further align representation across domains. These tasks utilize a shared feature encoder, effectively integrating their capabilities to improve the effectiveness of domain adaptation.} 
% The framework of our proposed method for unsupervised domain adaptation on point clouds. We initially train the model through supervised and self-supervised learning, then continue to optimize it via self-training. The framework comprises four critical components: translation distance prediction to alleviate centroid shift of point clouds, relational modeling to capture relationships between cross-domain samples, representation learning through supervised learning, and self-paced self-training to further align representation across domains. These tasks utilize a shared feature encoder, effectively integrating their capabilities to improve the effectiveness of domain adaptation. 
\label{framework}
\end{figure*}

\subsection{Self-supervised Learning of Point Clouds}
Self-supervised learning leverages the characteristics or intrinsic structure of the input data itself as the supervised signal to learn representations that contribute to downstream tasks by exploring the relationship or correlation between different input signals~\cite{poursaeed2020self}, \cite{sauder2019self}, \cite{gidaris2018unsupervised}, \cite{DBLP:conf/ijcai/JiaZZGZM23}, \cite{noroozi2017representation}, \cite{pathak2016context}. Recently, several works have applied self-supervised learning to point clouds. DefRec~\cite{DefRec} introduces the deformation-reconstruction task, while GAST~\cite{zou2021geometry} designs a deformation localization task based on it, simultaneously predicting the rotation angle of mixed point clouds. GLRV~\cite{GLRV} learns both global and local structures of point clouds by predicting the scaling factor and reconstructing compressed regions. ImplicitPCDA~\cite{GAI} incorporates learning geometry-aware implicit fields as a self-supervised task. MLSP~\cite{MLSP} encodes point clouds by predicting three distinct local attributes. 
In this paper, we propose a novel self-supervised learning method by regression on centroids' shift distance and relational learning with geometrically augmented samples, which can thus improve representations' quality of generalization and robustness.
% as a self-supervised task to understand the relationships between point clouds, which has not been considered in previous works.

\section{Methodology}
In the problem of unsupervised domain adaptation (UDA) on point cloud classification, given a labeled source domain 
$\mathcal{S}=\{(\boldsymbol{P}_{i}^{s},y_{i}^{s})\}_{i=1}^{n_{s}}$ and an unlabeled target domain 
$\mathcal{T}=\{(\boldsymbol{P}_i^t)\}_{i=1}^{n_t}$, where $P\in\mathbb{R}^{m\times3}$ represents a point cloud sample consisting of $m$ points, $y_i^s\in\mathcal{Y}=\{1,\cdots,C\}$, $C$ is the number of categories shared by the source and target domains. $n_s$ and $n_t$ are the number of point clouds in the two domains, respectively. 
We intend to train a deep neural network with point clouds from source and target domains, that can generalize well on unlabeled target point clouds. 
This is achieved by employing a two-part model: $\Phi = \Phi_{fea} \circ \Phi_{cls}$. The first part, $\Phi_{fea} : \mathbb{R}^{3} \rightarrow \mathbb{R}^{D} $, is a shared feature encoder that extracts representations of the input $\boldsymbol{P}$, with $D$ representing the feature dimension. The second part, $\Phi_{cls} : \mathbb{R}^{D} \rightarrow [0,1]^C $, is a classifier which maps $D$-dimensional feature vectors to a $C$-dimensional probability vector, indicating the likelihood of each of the $C$ classes. 
At the same time, self-supervised modules also constrain $\Phi_{fea}$ to facilitate model training. Our goal is to optimize this neural network for performance on $ \mathcal{Y} = \Phi(\boldsymbol{P})$. 

We use a typical unsupervised domain adaptation (UDA) framework to optimize $\Phi_{fea}$, and the overall pipeline is illustrated in Figure \ref{framework}. Specifically, our method mainly consists of three parts: a semantic classification task based on representation learning (see Section~\ref{Semantic}) and two self-supervised modules trained on the source and target domains, i.e, translation distance prediction to mitigate the suffering from shift of object centroids
(see Section~\ref{Translation}) and cascaded relational learning to improve robustness against topologically geometries' changes (see Section~\ref{Relational}).
% In addition to designing the pretext task of translation distance prediction, which plays a crucial role in learning characteristics of individual samples (see Section~\ref{Translation}), this paper also introduces the relational learning of multiple relation pairs to enable the model to understand the relationship between different samples (see Section~\ref{Relational}), so as to better improve the generalization ability of the model. We also introduce a novel combination method for data augmentation in Section~\ref{Relational}, which further improves the performance of relational learning. 
%In Section~\ref{Semantic}, we further align the source and target domains through self-paced self-training. Next, we will outline each proposed component and its training in detail.
During testing, an unseen sample from target domain can be fed into $\Phi = \Phi_{fea} \circ \Phi_{cls}$ directly to predict the probability of its semantic class, while the proposed regularization terms will not be utilized any more and thus cause no extra inference costs in comparison with its baseline SPST method \cite{zou2021geometry}.

\subsection{Self-Supervised Translation Augmentation}\label{Translation}

\begin{figure}[]
\centering
\includegraphics[height=2.0in,width=3.5in]{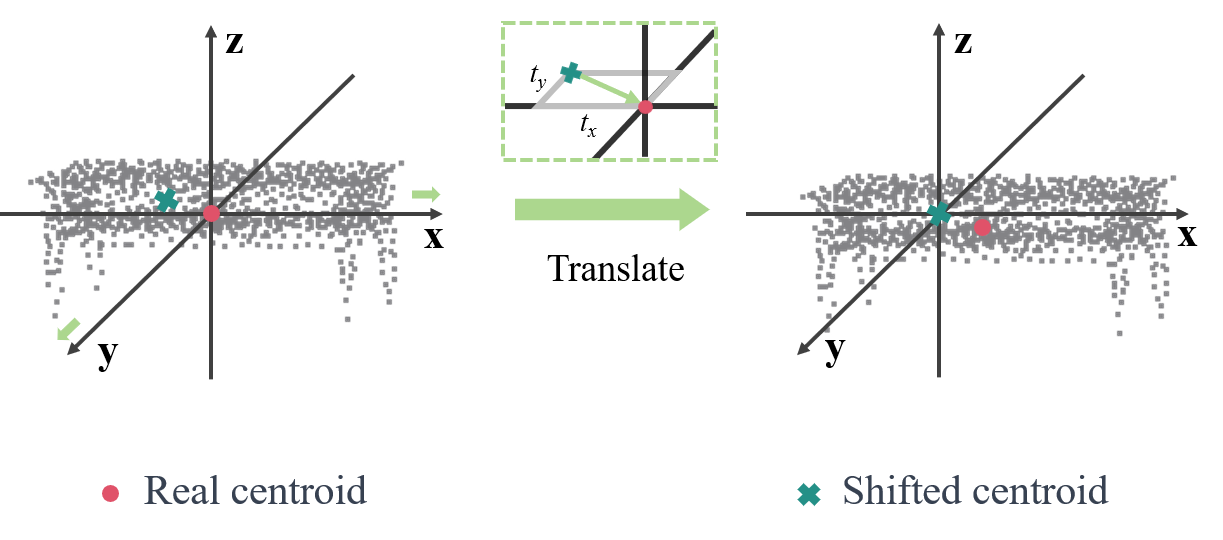}
\caption{The illustration of self-supervised translation augmentation. The sample is translated along the $x$-axis and $y$-axis, where the translation distance is determined by the maximum span of translation.} 
\label{trans}
\end{figure}

To improve generalization of point cloud representations on misalignment of object centroids caused by occlusion and noises, 
%enable the model to effectively deal with the issue of centroid shift due to the partial absence within a sample and to understand the characteristics of individual samples, 
we propose a translation distance prediction after 3D-to-2D projection of centroids on one plane that can encode simplified yet vital object pose (i.e. translation) into representations.   
%enables the model to have a global spatial awareness of the samples. 
Specifically, for one point cloud $\boldsymbol{P}$, we determine the translation distance $t$ based on the maximum span on the translation plane, e.g. the plane made up of the $x$ and $y$ axes in our scheme. 
Therefore, we can obtain the translation $t$ away from the origin on the projected translation plane given the input sample $\boldsymbol{P}_i$.
%is paper translates the input sample $P_i$ along the $x$-axis and $y$-axis to obtain its translation $P_i^t$ away from the origin. 
For example, considering $l_y$ denote the maximum span of $\boldsymbol{P}_i$ along the $y$-axis, the translation distance for the $i$-th point cloud are depicted by a set of pre-defined translation threshold $t_i \in \{t_i^1, t_i^2,t_i^3,t_i^4\}$, where the values of $t_i^j, j=1,2,\ldots,4$ increases sequentially. 
As the length of $l_y$ increases, so does the corresponding translation distance. 
In order to avoid excessively large translation distances that could lead to unstable convergence during training, we impose a cap on $t_i$, ensuring that $t_i \leq 0.1*l_y$. 
For each sample $\boldsymbol{P}_i^t$ after translation augmentation, we assign corresponding translation class labels $\bar{y}_i \in \{1,2,3,4\}$, according to the translation distance closest to those $t_i^j$. 
Following the feature encoder $\Phi_{fea}$, we integrate a distance classifier $\Phi_{trans}$. This arrangement allows us to compute the predicted translation probability vector $\hat{p_i} = \Phi_{trans}(\Phi_{fea}(\boldsymbol{P}_i^t))$. The loss function of translation distance prediction can be formulated as: 
\begin{equation}
\begin{split}
\mathcal{L}_{\mathrm{trans}}=-\frac1{n_s+n_t}\sum_{i=1}^{n_s+n_t}\sum_{t=1}^{T}(\mathbbm{1}[t=\bar{y}_{x,i}]\log\hat{p}_{i,t} \\ 
 +\mathbbm{1}[t=\bar{y}_{y,i}]\log\hat{p}_{i,t}),
\end{split}
\end{equation}
where $T=4$, $\hat{p}_{i,t}$ represents the $t$-th element of the translation prediction probability vector $\hat{p}_i$, and $\mathbbm{1}[\cdot]$ is an indicator function. The specifics of the translation process are illustrated in Figure \ref{trans}.

 \begin{figure*}[htbp]
\centering
\includegraphics[height= 3.8in,width=7.16in]{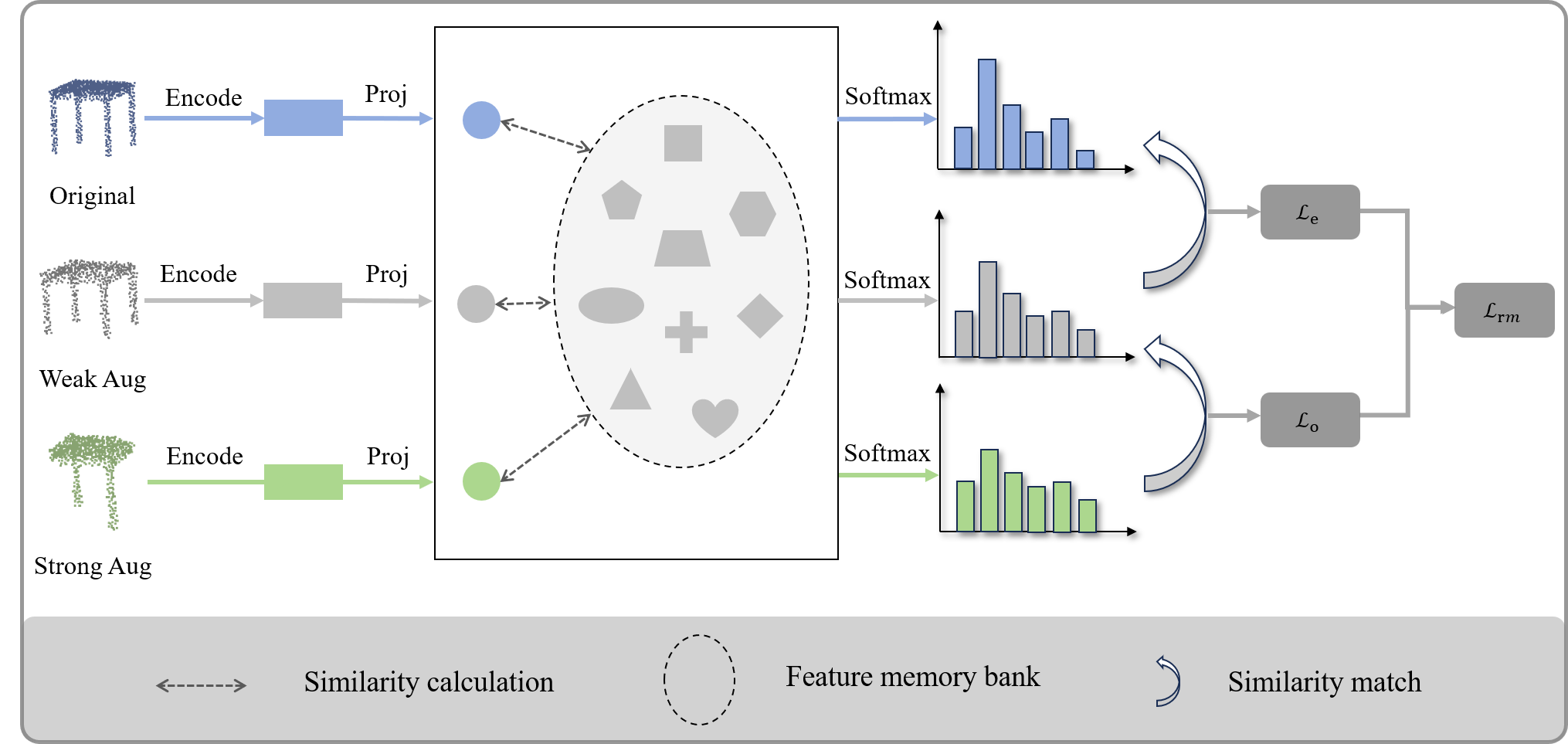} 
\caption{The illustration of relational learning with shape augmentation. Different augmented versions of the same sample are encoded and projected to the feature space, where similarities are calculated with features of other samples in the feature memory bank to derive the corresponding relational distribution. The relational distribution between two pairs are aligned to achieve relationship consistency.} 
\label{rlmp} 
\end{figure*}

\subsection{Relational Learning with Shape Augmentation}\label{Relational}

Relational learning encourages the model to utilize the relationships between different instances in representation space,  which are typically domain-invariant, thereby reducing the domain gap and improving the generalization ability of point cloud representations. 
However, existing relational learning methods often design the relationship constraints between augmented samples (e.g. weakly and strongly augmented samples), which can be further formulated into a cascade of relational self-supervised learning, by additionally incorporating the relationship between original samples and weakly augmented samples, as shown in Figure \ref{rlmp}. 
This strategy not only expands the model's decision boundary, enhancing its robustness against geometric variations, but also enables a more effective capture of intrinsic topological structure. Consequently, the proposed relational learning with shape augmentation can lead to a more uniform distribution of class centroids, further refining the model's performance. Due to the significant disparity between the original samples and the strongly augmented samples, making it challenging to discover their relations directly. Therefore, we constructed the original sample and weak augmentation, weak augmentation and strong augmentation as two pairs for relation learning, forming a gradual learning procedure.

For a given input point cloud $\boldsymbol{P}_i$, two augmented variants $\boldsymbol{P}_i^{wea} = T_w(\boldsymbol{P}_i)$ and $\boldsymbol{P}_i^{str} = T_s(\boldsymbol{P}_i)$ are obtained by data augmentation, and then the corresponding feature embeddings $\mathbf{\mathit{z}}_i = \Phi_{proj}(\Phi_{fea}(\boldsymbol{P}_i))$, $\mathbf{\mathit{z}}_i^w = \Phi_{proj}(\Phi_{fea}(\boldsymbol{P}_i^{wea}))$ and $\mathbf{\mathit{z}}_i^s = \Phi_{proj}(\Phi_{fea}(\boldsymbol{P}_i^{str}))$ are computed, where $T_w(\cdot)$ is a set of weak data augmentation methods, $T_s(\cdot)$ is a set of strong data augmentation methods, and $\Phi_{proj}(\cdot)$ is a projector used to project features into a feature space with uniform dimensions, facilitating the calculation of similarity distributions. 
Similar to the ReSSL~\cite{ressl}, we first calculate the similarity distribution of weakly augmented samples and strongly augmented samples with respect to the samples in the memory bank:
\begin{align}
\mathbf{r}_{ik}^w &= \frac{\exp(sim(\mathbf{\mathit{z}}_i^w,\mathbf{\mathit{z}}_k)/\tau_w)}{\sum_{j=1}^J\exp(sim(\mathbf{\mathit{z}}_i^w,\mathbf{\mathit{z}}_j)/\tau_w)},\\
\mathbf{r}_{ik}^s &= \frac{\exp(sim(\mathbf{\mathit{z}}_i^s,\mathbf{\mathit{z}}_k)/\tau_s)}{\sum_{j=1}^J\exp(sim(\mathbf{\mathit{z}}_i^s,\mathbf{\mathit{z}}_j)/\tau_s)},
\end{align}
where $ \tau_w$ and $\tau_s$ is the temperature coefficient, $\tau_w $\textless$ \tau_s$ to generate a sharper target distribution, $J = 65536$ is the number of samples in the memory bank, which dynamically maintains the most recent data $\{\mathbf{z}_k|k= 1,...,J\}$ using a FIFO method, followed by \cite{ressl} and $\mathbf{\mathit{z}}_k$ is the $k$-th sample among them. Then, we aim to maintain the consistency between the two similarity distributions through the cross-entropy loss function:
\begin{equation}
\mathcal{L}_{o} = H_{ce}(\mathbf{r}^w,\mathbf{r}^s).
\end{equation}

Similarly, in the relational self-supervised learning with the original and weakly augmented samples,
%we consider the input samples as weakly augmented samples, while the original weakly augmented samples can be relatively regarded as strongly augmented samples. 
the similarity distribution can be expressed as:
\begin{equation}
\mathbf{r}_{ik}=\frac{\exp(sim(\mathbf{\mathit{z}}_i,\mathbf{\mathit{z}}_k)/\tau)}{\sum_{j=1}^J\exp(sim(\mathbf{\mathit{z}}_i,\mathbf{\mathit{z}}_j)/\tau)},
\end{equation}
where $\tau $\textless$ \tau_w$ to keep the target distribution sharp. Then, the consistency of the relationship between the input samples and the weakly augmented samples can be guaranteed through supervising with the cross entropy loss:
\begin{equation}
\mathcal{L}_{e} = H_{ce}(\mathbf{r},\mathbf{r}^w).
\end{equation}
Note that $\mathbf{r}^w$ serves as the target distribution for calculating $\mathcal{L}_{o}$ and the online distribution for calculating $\mathcal{L}_{e}$, resulting in distinct values of $\tau_w$ for each case. Finally, the relational self-supervised learning for multiple relation pairs loss function we propose can be expressed as: 
\begin{equation}
\mathcal{L}_{rm} = \mathcal{L}_{o} +\lambda \mathcal{L}_{e},
\end{equation}
where $\lambda$ is a hyper-parameter that controls the importance of the additional relation terms.

\noindent\textbf{Data Augmentation --} Data augmentation plays an important role in relational learning, aiming to map samples to different views through random transformations, and the selection of data augmentation methods has a significant impact on the results~\cite{simslr}, \cite{moco}, \cite{byol}. 
In details, data augmentation can be leveraged to simulate various disturbances that might be encountered in the target domain. 
For instance, when point cloud samples from the target domain originate from the real world, they often have missing parts due to obstructions, which can be emulated using random cropping. 
Moreover, real-world point clouds frequently come with various types of noise, and random jittering can mimic the noise during data acquisition.
Evidently, integrating multiple data augmentation techniques can yield more discriminative feature representations. 
This paper presents a novel combination method that combines several commonly used data augmentation techniques in 3D vision, which can improve the diversity and difficulty of the augmented samples, and further improve the generalization ability of point cloud representations.

Specifically, for the input sample $\boldsymbol{P}_i$, we employ augmented methods with minor modifications (e.g., jittering) to obtain $\boldsymbol{P}_i^1 = \mathcal{T}_1(\boldsymbol{P}_i)$ and $\boldsymbol{P}_i^2 = \mathcal{T}_2(\boldsymbol{P}_i)$, followed by the farthest point sampling to obtain $\boldsymbol{P}_i^{f1} \in\mathbb{R}^{\left\lfloor \lambda\cdot m\right\rfloor\times3}$ and $\boldsymbol{P}_i^{f2} \in\mathbb{R}^{\left\lfloor (1-\lambda)\cdot m\right\rfloor\times3}$ from $\boldsymbol{P}_i^1$ and $\boldsymbol{P}_i^2$ respectively, where $\left\lfloor\lambda\cdot m \right\rfloor$ and $\left\lfloor (1-\lambda)\cdot m\right\rfloor$ are the numbers of points sampled from the point clouds, and $\left\lfloor\cdot\right\rfloor$ is rounded down. We then mix $\boldsymbol{P}_i^{f1}$ and $\boldsymbol{P}_i^{f2}$ to generate a new point cloud $\boldsymbol{P}_i^{m}$. 
Finally, we employ augmented methods with major modifications (e.g., cropping) to $\boldsymbol{P}_i^{m}$ to obtain a weakly augmented sample $\boldsymbol{P}_i^{wea} =\mathcal{T}_w(\boldsymbol{P}_i^{m})$ and a strongly augmented sample $\boldsymbol{P}_i^{str} =\mathcal{T}_s(\boldsymbol{P}_i^{m})$, where the augmentation methods used in $\mathcal{T}_s(\cdot)$ are more than those in $\mathcal{T}_w(\cdot)$.

\subsection{Semi-Supervised Representation Learning}\label{Semantic}

In context of UDA, the adopted self-training algorithm employ the labeled source data and unlabeled target data for domain adaptation, which can be in a manner of semi-supervised learning. 

\noindent\textbf{Supervised Learning --} We adopt a supervised learning strategy for the labeled source domain samples $\{(\boldsymbol{P}_{i}^{s},y_{i}^{s})\}_{i=1}^{n_{s}}$ , obtain the corresponding classification prediction 
$\{(\boldsymbol{p}_{i}^{s}\}_{i=1}^{n_{s}}$ through the feature extractor $\Phi_{fea}$ and classifier $\Phi_{cls}$, and optimize the model with the cross-entropy loss. 
%In order to make full use of the available information, we also take into account the weakly and strongly augmented samples generated in relational learning. 
Since the augmented samples generated in relational learning share the same class labels with the original samples, the loss function can be depicted as follows: 
\begin{equation}
\begin{split}
\mathcal{L}_{\text{cls}}^s=-\frac{1}{n_s}\sum_{i=1}^{n_s}\sum_{c=1}^C\mathbbm{1}[c=y_i^s] 
(\log p_{i,c}^s \\
+ \log p_{i,c}^{sw} + \log p_{i,c}^{ss}),
\end{split}
\end{equation}
where $p_{i,c}^s$ represents the $c$-th element of the classification prediction probability ${p}_{i}^{s} = \Phi_{cls}(\Phi_{fea}(P_i^s))$. $p_{i}^{sw}$ and $p_{i}^{ss}$ represent the classification predictions of weakly and strongly augmented samples, respectively.

\vspace{0.1cm} \noindent\textbf{Self-Paced Self-Training --} In addition to using self-supervised learning to reduce domain gap, we also employ the popular self-training method to further boost the accuracy of domain adaptation. 
%The effectiveness of the self-training method is closely related to the reliability of the pseudo-labels. 
Inspired by the GAST~\cite{zou2021geometry}, we adopt a self-paced learning strategy to select reliable samples from target domain for assigning pseudo labels. 
We first designate the category with the maximum prediction probability for a target sample as its pseudo-label. Only when this prediction probability exceeds the specified threshold, such a target sample will be adopted for self-training. Finally, the self-training loss function is as follows:
\begin{equation}
\mathcal{L}_{\text{cls}}^t = - \frac 1 { n_t }\sum_{i=1}^{n_t}\left(\sum_{c=1}^C\hat{y}_{i,c}^t\log p_{i,c}^t+\gamma|\hat{y}_i^t|_1\right).\label{eqn.spst}
\end{equation}
The first term in Eqn (\ref{eqn.spst}) calculates the cross entropy between the prediction and the pseudo-label, aiming to optimize the semantic classifier. 
The objective of the second term is to prevent degenerate solutions, where the prediction probability of all pseudo label corresponding categories is less than the threshold, resulting in omitting in the following refining stage.

\begin{table*}[t]
\caption{Comparative evaluation in classification accuracy ($\%$) averaged over 3 seeds ($\pm$ SEM) on the PointDA-10 dataset. The best results in each column are in bold}
\renewcommand\arraystretch{1}
\centering
\begin{tabular}{c|ccccccc}
\hline
  Method    & M $\rightarrow$ S      & M $\rightarrow$ $S^*$        & S $\rightarrow$ M   & S $\rightarrow$ $S^*$              & $S^*$ $\rightarrow$ M              & $S^*$ $\rightarrow$ S              & Avg.            
  \\ 
  \hline

   Supervised    & 93.9 $\pm$ 0.2       & 78.4 $\pm$ 0.6          & 96.2 $\pm$ 0.1          & 78.4 $\pm$ 0.6         & 96.2 $\pm$ 0.1         & 93.9 $\pm$ 0.2         & 89.5 $\pm$ 0.3      \\

    w/o Adapt        &   83.3 $\pm$ 0.7    &  43.8 $\pm$ 2.3     &   75.5 $\pm$ 1.8 &    42.5 $\pm$ 1.4     &   63.8 $\pm$ 3.9    &   64.2 $\pm$ 0.8    &    62.2 $\pm$ 1.8   \\ \hline

    DANN~\cite{dann}   & 74.8 $\pm$ 2.8 &    42.1 $\pm$ 0.6   & 57.5 $\pm$ 0.4 & 50.9 $\pm$ 1.0 &   43.7 $\pm$ 2.9 &   71.6 $\pm$ 1.0  &   56.8 $\pm$ 1.5 \\
    
    PointDAN~\cite{pointdan}        & 83.9 $\pm$ 0.3 &    44.8 $\pm$ 1.4   & 63.3 $\pm$ 1.1 & 45.7 $\pm$ 0.7 &   43.6 $\pm$ 2.0 &   56.4 $\pm$ 1.5  &  56.3 $\pm$ 1.2 \\

    RS~\cite{sauder2019self}      & 79.9 $\pm$ 0.8 &    46.7 $\pm$ 4.8   & 75.2 $\pm$ 2.0 & 51.4 $\pm$ 3.9 &   71.8 $\pm$ 2.3 &   71.2 $\pm$ 2.8  &   66.0 $\pm$ 1.6 \\

    DefRec + PCM ~\cite{DefRec}   & 81.7 $\pm$ 0.6 &    51.8 $\pm$ 0.3  & 78.6 $\pm$ 0.7 & 54.5 $\pm$ 0.3 &   73.7 $\pm$ 1.6 &   71.1 $\pm$ 1.4  &   68.6 $\pm$ 0.8 \\

     GAST ~\cite{zou2021geometry}     & 84.8 $\pm$ 0.1 & 59.8 $\pm$ 0.2 & 80.8 $\pm$ 0.6 & 56.7 $\pm$ 0.2 & 81.1 $\pm$ 0.8 & 74.9 $\pm$ 0.5 & 73.0 $\pm$ 0.4 \\

     GLRV~\cite{GLRV}         &   85.4 $\pm$ 0.4     &    \textbf{60.4} $\pm$ 0.4      &   78.8 $\pm$ 0.6  &     \textbf{57.7} $\pm$ 0.4      &     77.8 $\pm$ 1.1    &      76.2 $\pm$ 0.6     &   72.7 $\pm$ 0.6     \\ 
     
     ImplicitPCDA ~\cite{GAI}   &   86.2 $\pm$ 0.2     &    58.6 $\pm$ 0.1      &   81.4 $\pm$ 0.4  &     56.9 $\pm$ 0.2      &    81.5 $\pm$ 0.5    &    74.4 $\pm$ 0.6    &    73.2 $\pm$ 0.3   \\

     MLSP ~\cite{MLSP} &   85.7 $\pm$ 0.6     &   59.4 $\pm$ 1.3      &   82.3 $\pm$ 0.9  &     57.3 $\pm$ 0.7      &    82.2 $\pm$ 0.5    &    76.4 $\pm$ 0.5    &    73.8 $\pm$ 1.0   \\

     % Self-dist$\dag$ ~\cite{Self-dist}&  83.9 $\pm$ 0.0    &   \textbf{61.1}  $\pm$ 0.0     &   80.3 $\pm$ 0.0  &     \textbf{58.9} $\pm$ 0.0     &   \textbf{85.5} $\pm$ 0.0  &   80.9  $\pm$ 0.0  &   75.1 $\pm$ 0.0  \\ 

     COT ~\cite{COT} &   84.7 $\pm$ 0.2     &   57.6 $\pm$ 0.2      &   \textbf{89.6} $\pm$ 1.4  &     51.6 $\pm$ 0.8      &    \textbf{85.5} $\pm$ 2.2    &    77.6 $\pm$ 0.5    &   74.4 $\pm$ 0.9   \\

     Ours  &   \textbf{86.5} $\pm$ 0.3    & 59.5 $\pm$ 0.1   &   85.2 $\pm$ 0.5 &   57.4 $\pm$ 0.1     &  82.4 $\pm$ 0.5   &   \textbf{81.5} $\pm$ 0.7  &  \textbf{75.4} $\pm$ 0.1 \\ 
     
     \hline

\end{tabular}

\label{compare}
\end{table*}

\begin{table*}[t]
\caption{Ablation study on the PointDA-10 dataset. "*" indicates that the method does not employ self-training. The best results in each column are in bold}
\centering

\renewcommand\arraystretch{1}
\centering
\begin{tabular}{ccc|ccccccc}

\hline

    &  Trans & RL & M $\rightarrow$ S      & M $\rightarrow$ $S^*$        & S $\rightarrow$ M   & S $\rightarrow$ $S^*$        & $S^*$ $\rightarrow$ M              & $S^*$ $\rightarrow$ S        & Avg.           \\ 
    
      \hline

    RS ~\cite{sauder2019self}  &      &      & 79.9  &    46.7   & 75.2& 51.4  &   71.8  &   71.2   &   66.0  \\

    DefRec + PCM ~\cite{DefRec} &      &    & 81.7  &    51.8  & 78.6 & 54.5 &   73.7 &   71.1  &   68.6  \\
    
    GAST*~\cite{zou2021geometry}&  &          & 83.9         & 56.7         & 76.4          & 55.0        & 73.4      & 72.2         & 69.5         \\

    ImplicitPCDA*~\cite{GAI}&    &        &   85.8     &    55.3      &   77.2  &     55.4       &    73.8    &    72.4    &    70.0   \\

    MLSP*   ~\cite{MLSP}     &    &        &   83.7      &   55.4     &   77.1   &     \textbf{55.6}     &    78.2    &    76.1   &    71.0    \\

    COT* ~\cite{COT} &    &        &  83.2     &   54.6    &   78.5   &     53.3     &   \textbf{79.4}    &   \textbf{77.4}   &    71.0    \\

      \hline

   \multirow{3}{*}{Ours*} &  $\checkmark$     &      &  84.0     &   55.6    &   79.0  &   51.6   &    74.3   &   69.1 &   68.9  \\ 

     &  &     $\checkmark$       & 83.8    & 55.8    & 78.9       & 53.3      & 75.6     & 72.0     & 69.9     \\

   & $\checkmark$       & $\checkmark$         & \textbf{84.1} &  \textbf{57.6}   &  \textbf{81.5} & 55.0   &  78.2 &   74.7  &  \textbf{71.8} \\  
       \hline
\end{tabular}

\label{ablation}
\end{table*}

\subsection{Overall Training}
The overall loss of our method includes two self-supervised losses and two classification losses:
\begin{equation}
\mathcal{L} = \mathcal{L}_{rm} + \alpha \mathcal{L}_{trans} + 
\beta \mathcal{L}_{cls}^s + \eta \mathcal{L}_{cls}^t,
\end{equation}
where $\alpha$, $\beta$ and $\eta$ are hyper-parameters used to balance the weights between methods. 
Note that, during the early stages of model training, we mainly rely on the first three loss terms to ensure better completion of the adaptation process. 
Once the initial training is completed, we use the model to generate pseudo-labels for the target domain samples and proceed with self-training.

\section{Experiments}

\subsection{Datasets}
PointDA-10~\cite{pointdan} is a widely used dataset for point cloud domain adaptation, which consists of three subsets: ModelNet40~\cite{modelnet}, ShapeNet~\cite{shapenet} and ScanNet~\cite{scannet}. $10$ common categories (sofa, lamp, chair, etc.) in these three datasets are chosen for experiments, named ModelNet-10 (\textit{M}), ShapeNet-10 (\textit{S}) and ScanNet-10 (\textit{S*}). \textit{M} and \textit{S} are both synthetic point cloud datasets sampled from CAD models, where \textit{M} contains $4,183$ training samples and $856$ test samples, and \textit{S} contains $17,378$ training samples and $2,492$ test samples. Point clouds in \textit{S*} are collected  from real-world indoor scenes, containing $6,110$ training samples and $2,048$ test samples, which are incomplete due to the occlusion of surrounding objects.

\subsection{Implementation}
In our experiments, DGCNN~\cite{dynamic} is used as the feature extractor. The Adam optimizer~\cite{kingma2014adam} is used with an initial learning rate of $0.001$ and a weight decay of $0.00005$, along with the application of an epoch-wise cosine annealing learning rate scheduler. We train all methods for $200$ epochs using three different random seeds with a batch size of $32$ on an NVIDIA GTX $4090$ GPU.

\begin{table*}[t]
\caption{The impact of translation dimensions. The best results in each column are in bold}
\centering
\begin{tabular}{cccc|ccccccc}
\hline
  & X  & Y & Z  & M $\rightarrow$ S      & M $\rightarrow$ $S^*$        & S $\rightarrow$ M   & S $\rightarrow$ $S^*$              & $S^*$ $\rightarrow$ M              & $S^*$ $\rightarrow$ S              & Avg.              \\ \hline

    \multirow{4}{*}{Trans}  &  \checkmark     &  \checkmark   &        & \textbf{84.0}   &   \textbf{55.6}     &  \textbf{79.0}  &     \textbf{51.6}    &   74.3  &  \textbf{69.1}     &   \textbf{68.9}  \\

    &     \checkmark  &     &     \checkmark   &  83.9    &   53.3     &  78.7  &    48.6     &   \textbf{75.1} &   65.4    &    67.5  \\

    &       &  \checkmark   &    \checkmark    &  83.1   &    53.0    & 78.5   &    48.4   &   74.8  &    \textbf{69.1}   &    67.8  \\

    &    \checkmark   &  \checkmark   &    \checkmark    &  83.6   &      54.4  &  78.3  &   51.3      &   73.9  &    67.9   &  68.2    \\

     \hline

\end{tabular}

\label{dim}
\end{table*}

\begin{table*}[t]
\caption{The impact of weak and strong augmentation methods on relational learning. "$J$", "$S$", and "$C$" denote jittering, scaling, and cropping, respectively. where "$C_w$" retains more points than "$C_s$" in the cropping operation. The best results in each column are in bold}
\centering
\begin{tabular}{cc|ccccccc}
\hline
  WA & SA & M $\rightarrow$ S      & M $\rightarrow$ $S^*$        & S $\rightarrow$ M   & S $\rightarrow$ $S^*$              & $S^*$ $\rightarrow$ M              & $S^*$ $\rightarrow$ S              & Avg.              \\ 
  \hline

      J   &   JS    & 81.7   &   51.7     &  78.4  &    47.0    &   70.4  &  63.5     &   65.5 \\

       J$C_w$  &   J$C_s$    &  83.5   &  54.8     & 77.6  &    52.0    &   74.2 &   70.0    &   68.7 \\

      J$C_w$   &    J$C_s$S   &  \textbf{83.8}   &    \textbf{55.8}  & \textbf{78.9}  &   \textbf{53.3}   &   \textbf{75.6}  &    \textbf{72.0}   &   \textbf{69.9} \\
       
     \hline

\end{tabular}

\label{ws}
\end{table*}

\subsection{Comparison with the State-of-the-art Methods}
Our method is compared with the state-of-the-art point cloud domain adaptation methods on the PointDA-10 dataset, including Domain Adversarial Neural Network (DANN)~\cite{dann}, Point Domain Adaptation Network (PointDAN)~\cite{pointdan}, Reconstruction Space Network (RS)~\cite{sauder2019self}, Deformation Reconstruction Network with Point Cloud Mixup (DefRec+PCM)~\cite{DefRec}, Geometry-aware self-training (GAST)~\cite{zou2021geometry}, Global-Local Structure Modeling with Reliable Voted Pseudo Labels (GLRV)~\cite{GLRV}, Geometry-Aware Implicits (ImplicitPCDA)~\cite{GAI}, Masked Local Structure Prediction (MLSP)~\cite{MLSP}, and Contrastive Learning and Optimal Transport (COT)~\cite{COT}. The supervised method trains the model using only labeled target samples, while the w/o adapt method uses only labeled source samples. 
% Self-Distillation (Self-dist)~\cite{Self-dist} 

As shown in Table \ref{compare}, our method achieves the best results in both experimental settings $M \rightarrow S$ and $S^* \rightarrow S$, and all the results are consistently ranked among top 2. Additionally, we obtain state-of-the-art results on average accuracy, surpassing the previously best method COT by $1\%$, with an obvious increase in $M \rightarrow S^*$ and $S \rightarrow S^*$ by $1.9\%$ and $5.8\%$. Compared with DefRec+PCM, which is the first work to employ self-supervision in point cloud UDA, our method exhibits a $6.8\%$ improvement. It also achieves increases of $2.4\%$, $2.7\%$, $ 2.2\%$ and $ 1.6\%$ against GAST, GLRV, ImplicitPCDA and MLSP, respectively. %While our method shows a modest improvement of $0.3\%$ in average accuracy over Self-dist, it demonstrates a more significant advantage in the experimental settings  $M \rightarrow S$ and  $S \rightarrow M$, outperforming by $2.6\%$ and $4.9\%$. 
Considering the superiority of Optimal Transport ($OT$) in the COT to the conventional self-paced self-training in our method, the performance gap between both methods can be larger and credited to the proposed self-supervised geometric augmentation, which can further verify the effectiveness of our method.
% These results highlight the effectiveness of integrating relational learning, which explores relationships between different samples, in enhancing the model's generalization abilities. 

\begin{table*}[t]
\caption{The impact of our proposed data augmentation method}
\renewcommand\arraystretch{1}
\centering
\begin{tabular}{c|ccccccc}
\hline
  Method    & M $\rightarrow$ S      & M $\rightarrow$ $S^*$      & S $\rightarrow$ M   & S $\rightarrow$ $S^*$        & $S^*$ $\rightarrow$ M          & $S^*$ $\rightarrow$ S     & Avg.             \\ 
  \hline

     RL &  \textbf{83.8}   &  \textbf{55.8}     &  \textbf{78.9}  &    \textbf{53.3}      &    \textbf{75.6}   &   \textbf{72.0}   &   \textbf{69.9}   \\

    w/o Aug &  83.6   &  55.2  &  77.8 &   53.1    &  73.5   &   69.4 &   68.8   \\ 
     \hline
\end{tabular}
\label{da}
\end{table*}

\begin{figure}[!t]
\centering
\subfloat[]{\includegraphics[width=1.69in]{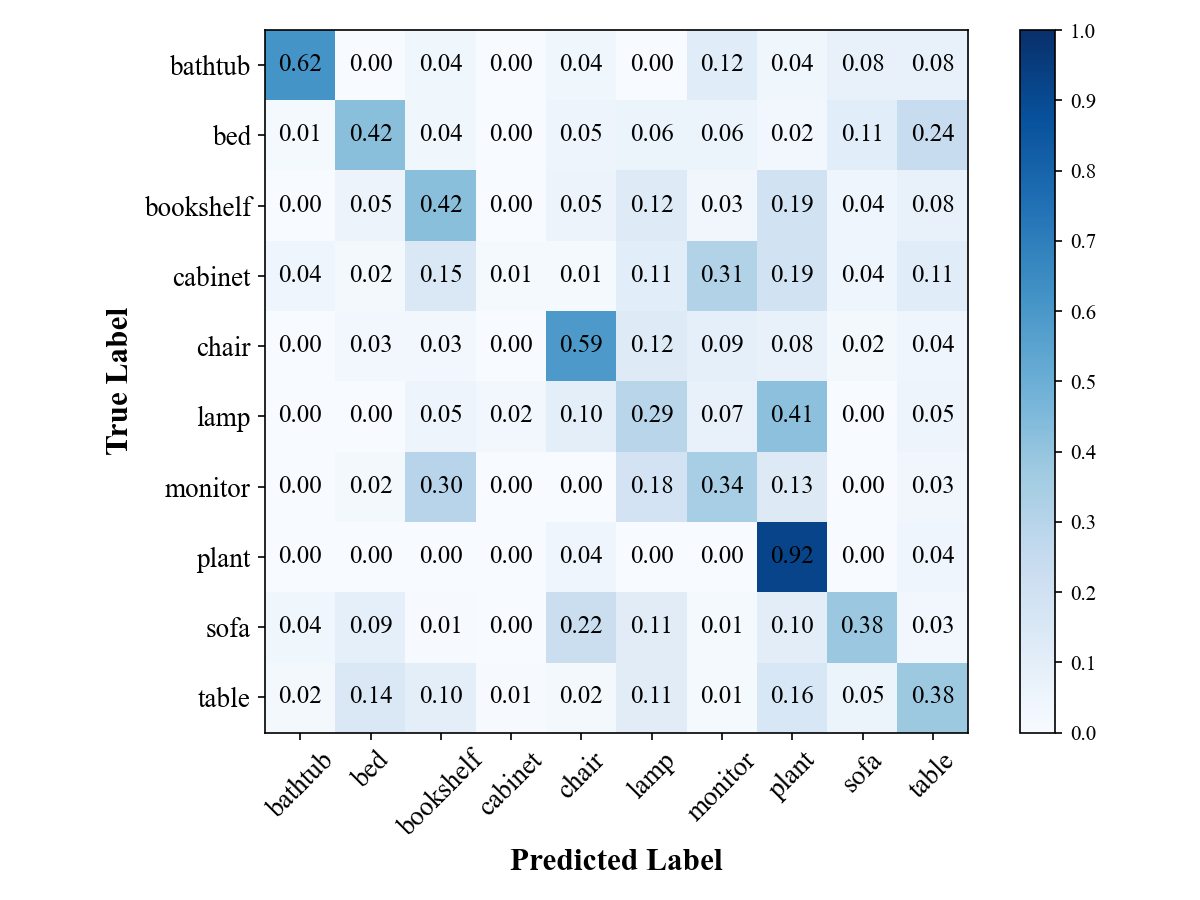}
\label{fig:subfig1m}}
\hfil
\subfloat[]{\includegraphics[width=1.69in]{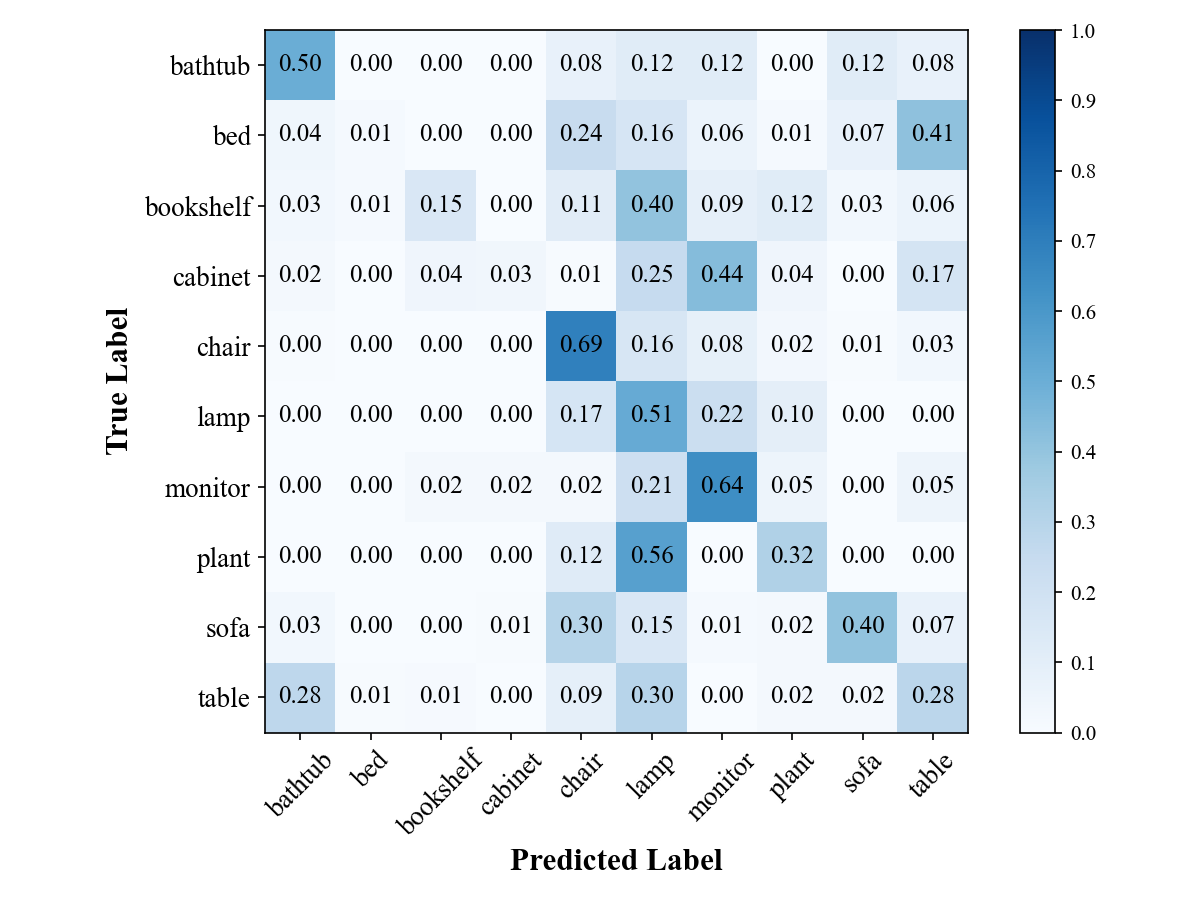}
\label{fig:subfig2m}}

\subfloat[]{\includegraphics[width=1.69in]{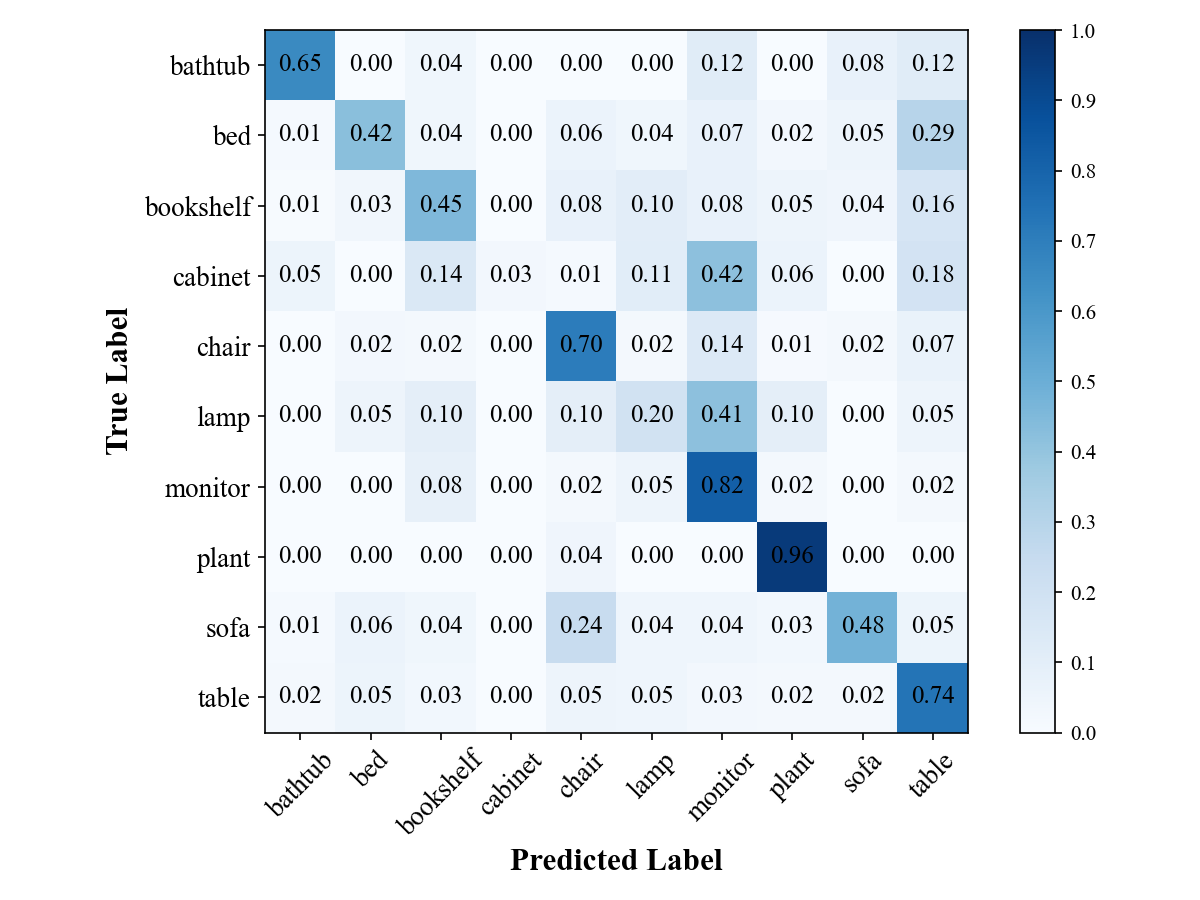}
\label{fig:subfig3m}}
\hfil
\subfloat[]{\includegraphics[width=1.69in]{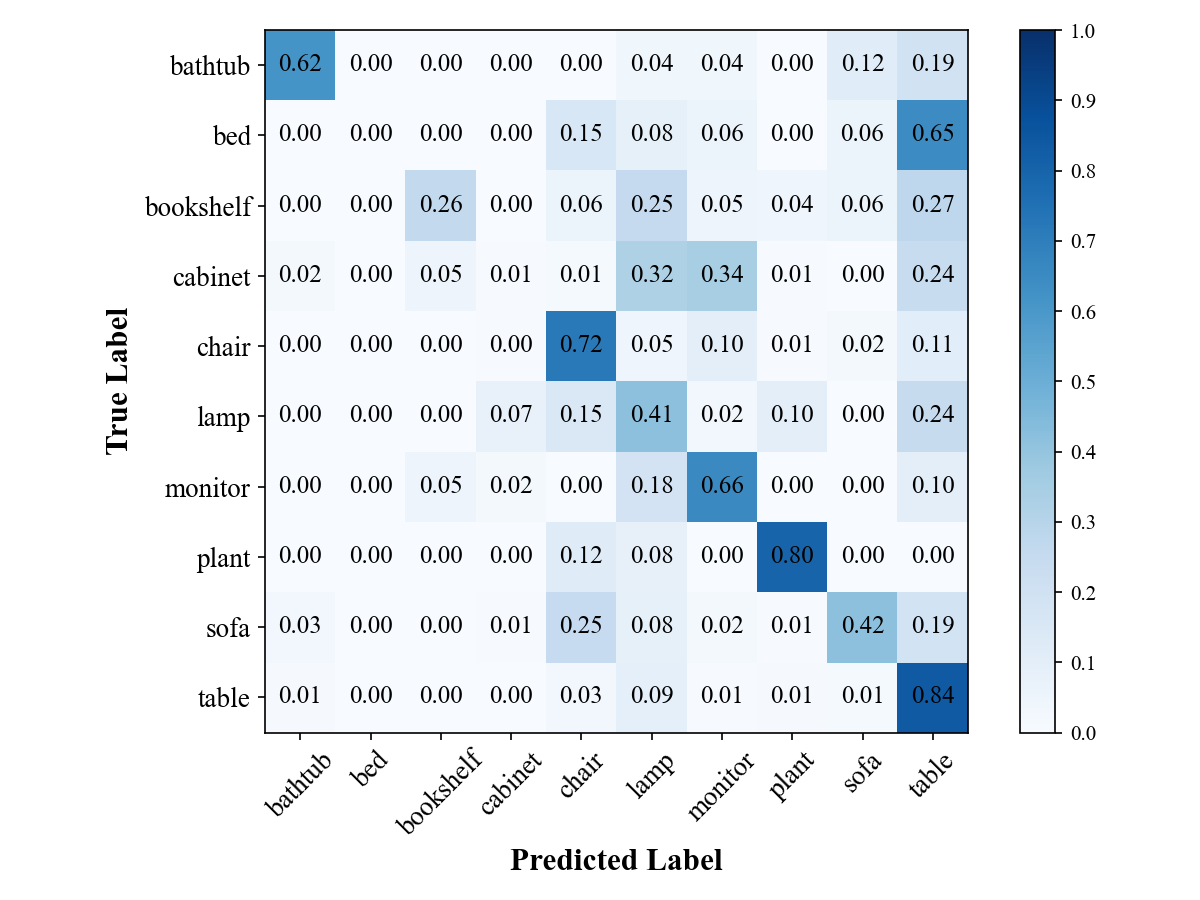}
\label{fig:subfig4m}}

\caption{(a) w/o Adapt: M $\rightarrow$ S*. (b) w/o Adapt: S $\rightarrow$ S*. (c) w/ Adapt: M $\rightarrow$ S*. (d) w/ Adapt: S $\rightarrow$ S*. Confusion matrices for the classification of test samples on the target domain. Darker colors within the visualization reflect higher levels of accuracy.}
\label{cm}
\end{figure}

\subsection{Ablation Study}
\noindent\textbf{The impact of translation distance prediction and relational self-supervised learning}.
To investigate the effectiveness of the two proposed self-supervised methods, we conduct an ablation study on six transfer scenarios on the PointDA-10 dataset, and the results are shown in Table \ref{ablation}. Both methods have a positive impact on the results, and their joint application further improves the results. This indicates that combining these two methods not only helps the model better understand the characteristics of individual samples, but also helps the model learn the complex relationships between samples, thereby achieving better performance in transfer and generalization between different domains. Specifically, in the $M \rightarrow S^*$ and $S \rightarrow S^*$ experimental settings, using only relational learning achieves performances of $55.8\%$ and $53.3\%$, respectively. With the addition of the translation distance prediction task, the performance improves by $1.8\%$ and $1.7\%$, respectively. These results prove the effectiveness of our proposed self-supervised task in alleviating the issue of centroid shift in point clouds in real-world scenarios. Compared to previous work, this paper achieves the best results in three out of six scenarios using only self-supervised methods. The average accuracy is $0.8\%$ higher than the previously best-performing MLSP and COT.

\begin{figure}[!t]
\centering
\subfloat[]{\includegraphics[width=1.69in]{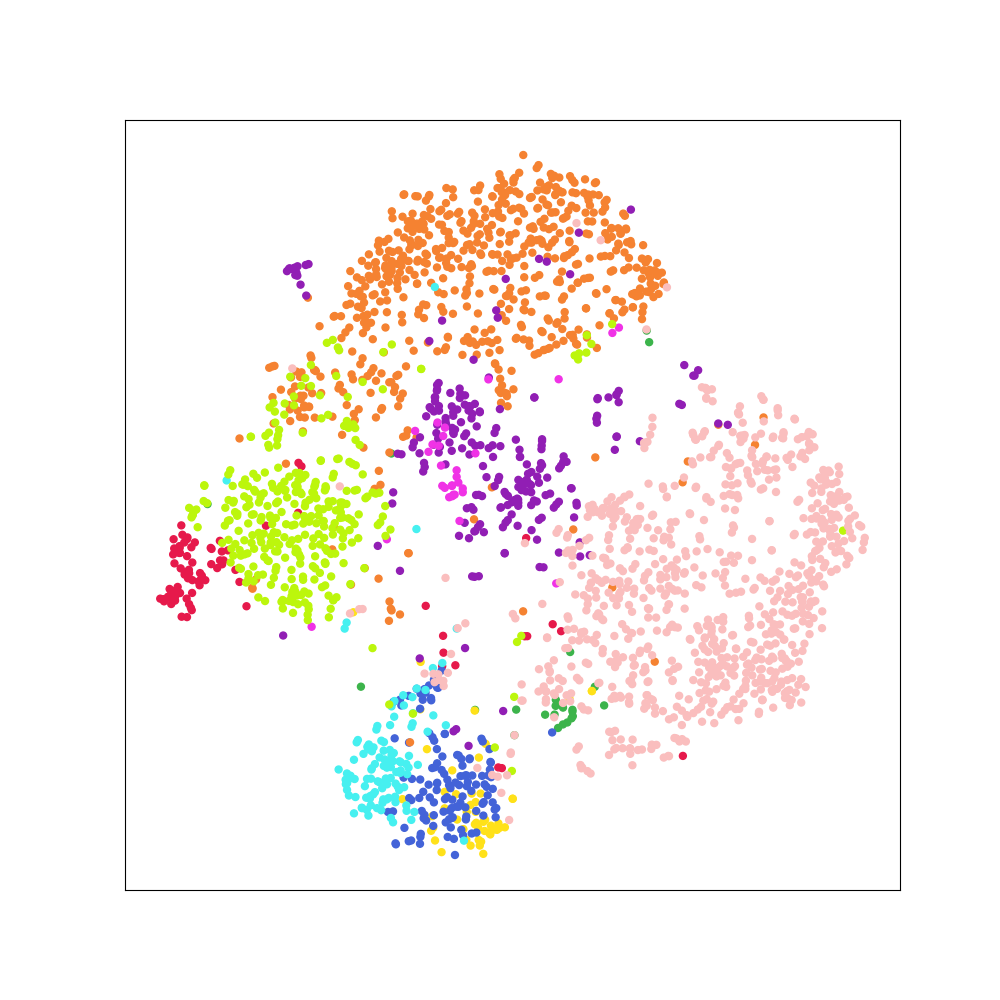}
\label{fig:subfig1}}
\hfil
\subfloat[]{\includegraphics[width=1.69in]{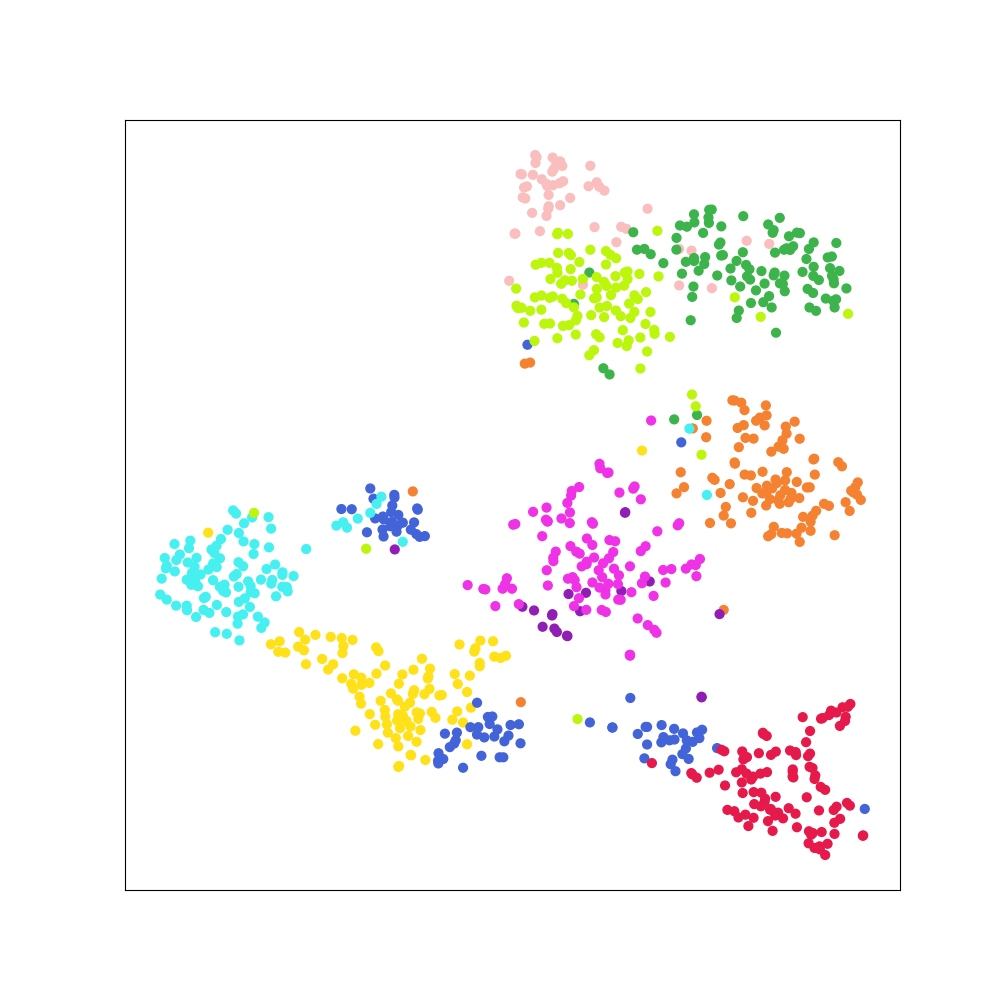}
\label{fig:subfig2}}

\subfloat[]{\includegraphics[width=1.69in]{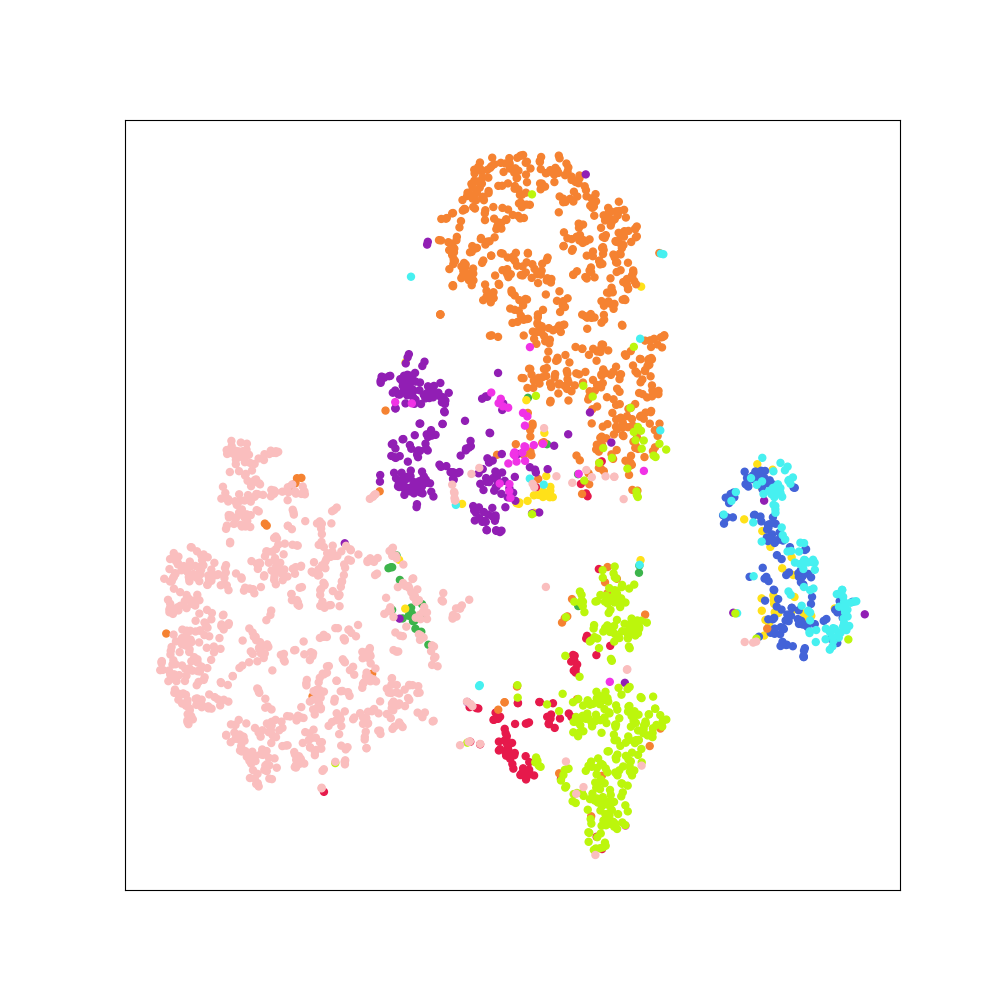}
\label{fig:subfig3}}
\hfil
\subfloat[]{\includegraphics[width=1.69in]{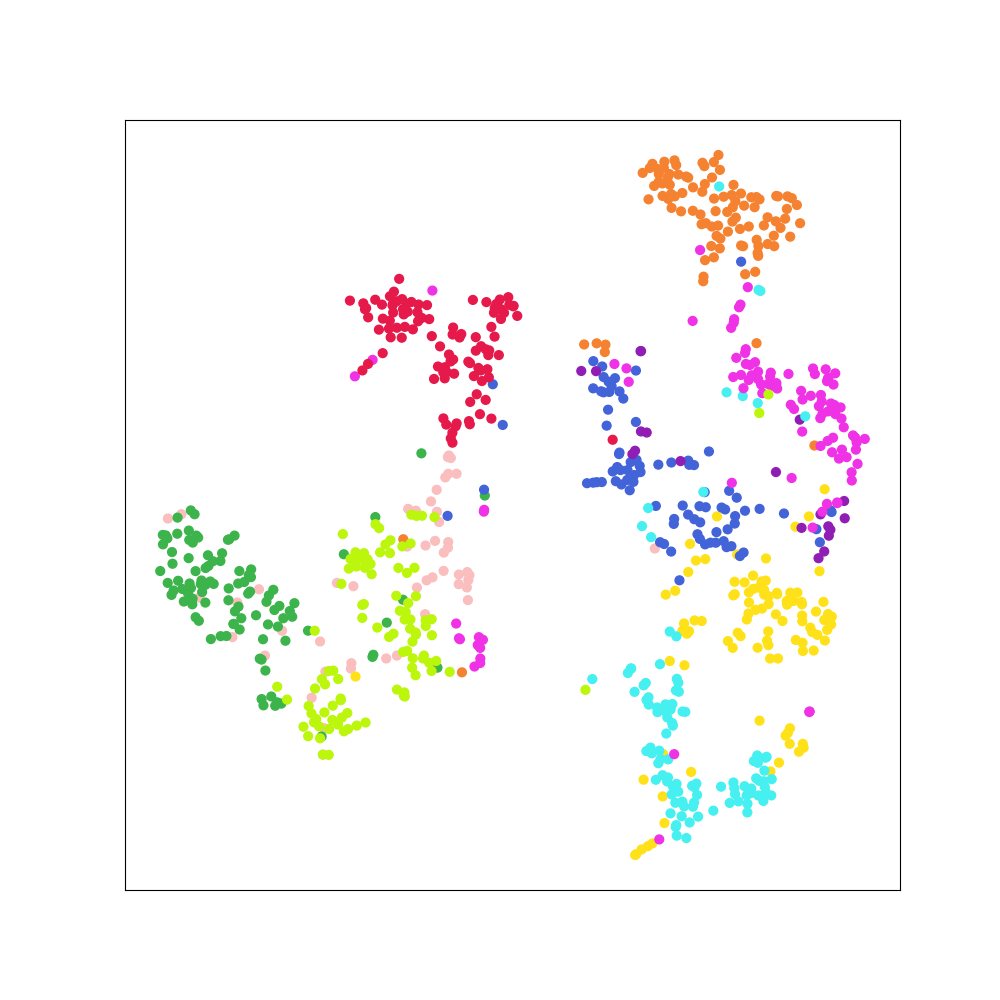}
\label{fig:subfig4}}

\caption{(a) w/o Adapt: S* $\rightarrow$ S. (b) w/o Adapt: S* $\rightarrow$ M. (c) w/ Adapt: S* $\rightarrow$ S. (d) w/ Adapt: S* $\rightarrow$ M. The t-SNE visualization of feature distribution on the target domain. Different colors represent different classes.}
\label{fig:main}
\end{figure}

\noindent\textbf{The impact of translation dimensions}. 
To investigate the impact of different translation dimensions on the pretext task of predicting translation distances, we analyze the performance differences when translations are made along different combinations of coordinate axes. The results as shown in Table \ref{dim} indicate that configurations involving translations exclusively in the horizontal dimensions achieve the highest average performance ($68.9\%$). In contrast, when the translation involves the vertical dimension, there is a notable drop in classification accuracy, decreasing by $1.4\%$ and $1.1\%$, respectively. Such results can be explained by the larger span of the point cloud collected from the real world on the horizontal plane (consisting of $X$ and $Y$ axes) compared to the smaller span along the $Z$-axis.
For example, given point clouds of the sofa and bed classes, when the point clouds miss parts, the displacement of the center points on the $X$ and $Y$ axes will be greater. 
Additionally, when translations encompass all three spatial dimensions, there is a reduction of $0.7\%$ compared to considering only the horizontal dimensions. This indicates that the complexities introduced by vertical translations might adversely affect the model's predictive performance, hence our approach is to predict translation distances on the horizontal dimensions.

\noindent\textbf{The impact of weak and strong augmentation methods}. 
To explore the effects of weak and strong augmentation on relational learning, different combinations of augmentation methods are used to perform both weak and strong augmentations. We use jittering, random cropping (retaining 50\%-80\% of the points), and scaling for strong augmentation, while employ jittering and random cropping (retaining 60\%-90\% of the points) for weak augmentation. The results from Table \ref{ws} indicate that random cropping significantly impacts the performance of the model. Removing the cropping operation resulted in a $4.4\%$ decrease in average performance, demonstrating that cropping effectively simulates the missing point cloud issues caused by occlusions in the real world. Additionally, the performance further improved by $1.2\%$ after introducing scaling, which effectively reduces the domain gap between synthetic and real point clouds by aligning their scale and density, thereby enhancing the model's generalization capabilities.

\noindent\textbf{The impact of data augmentation}. 
To verify the impact of our proposed data augmentation method in relational learning, an ablation study is performed on the PointDA-10 dataset, where w/o Aug uses a simple series of augmentation methods (e.g., jittering, cropping, etc.). As shown in Table \ref{da}, gains of $2.1\%$ and $2.6\%$ are achieved in $ S^* \rightarrow M $ and $ S^* \rightarrow S $, respectively. This suggests that the proposed augmentation method effectively enhances the model's comprehension of real point clouds, thus further improving the experimental results when real point clouds datasets are used as the source domain.

\noindent\textbf{Class-Wise Accuracy Visualization}.
We utilize confusion matrices to visualize the predictive accuracy of our model across different categories, where rows represent the actual categories and columns represent the predicted ones. This approach not only displays the overall accuracy of the model, but also highlights the categories that are prone to classification errors. as shown in Fig. \ref{cm}, visualization of confusion matrices illustrating class-wise classification accuracy for the baseline (w/o Adapt) and our method (w/ Adapt) on $M \rightarrow S^*$ and $S \rightarrow S^*$. Fig. \ref{fig:subfig1m} and \ref{fig:subfig2m} show the results without adaptation, whereas Fig. \ref{fig:subfig3m} and \ref{fig:subfig4m} display the results with adaptation. The visualization reveals that the diagonal lines in the confusion matrices for Fig. \ref{fig:subfig3m} and \ref{fig:subfig4m} are darker, indicating a higher overall accuracy. Additionally, the colors in the upper and lower triangles are comparatively lighter, suggesting that fewer categories are confused. This demonstrates that our proposed method effectively reduces the domain gap between the source and target domains, thereby enhancing the model's accuracy in recognizing samples in the target domain.

\noindent\textbf{Feature Visualization}. 
We use t-SNE~\cite{van2008visualizing} to visualize the feature distribution on the target domains of the UDA tasks $ S^* \rightarrow S $ and $ S^* \rightarrow M $ for both the baseline and our proposed method in Fig. \ref{fig:main}. Fig. \ref{fig:subfig1} and \ref{fig:subfig2} display the feature distributions obtained without using adaptive methods, whereas Fig. \ref{fig:subfig3} and \ref{fig:subfig4} are the feature distributions obtained using the adaptive methods proposed in this paper. Without domain adaptation, the features of different classes in the target domain tend to overlap. With domain adaptation, the feature distribution in the target domain begins to converge, resulting in clear clustering and effectively reducing the mingling of features from different classes.

\section{Conclusion}
In this paper, we propose a novel point cloud representation learning via self-supervised geometric augmentation, aiming to narrow the gap between the synthetic source domain and the real-world target domain. On the one hand, a translation distance prediction pretext task is designed to mimic the centroid shift of point clouds due to occlusion and noise. On the other hand, a cascaded relational self-supervised learning on geometrically-augmented point clouds is introduced for the first time in 3D UDA as constraints of cross-domain geometric features. Extensive experimental results demonstate that our approach is superior to existing methods. Although the proposed method can effectively close the domain gap with self-supervised geometric augmentation, relation learning in our scheme is sensitive to data augmentation techniques, which potentially limits its flexibility. This sensitivity to augmentation techniques is an issue that needs to be further addressed in future work. 
%In this paper, we introduce relational learning into point clouds UDA for the first time, and propose a self-supervised task for translation distance prediction. The combination of these two strategies not only enhances the model's understanding of individual sample characteristics, but also significantly improves the model's ability to generalize across different domains by learning the complex relationships between samples, effectively mitigating the impact caused by the domain gap. The effectiveness of our approach is validated on the PointDA-10 benchmark, achieving state-of-the-art performance.

\bibliographystyle{IEEEtran}
\bibliography{paper}

\newpage

\begin{IEEEbiography}[{\includegraphics[width=1in,height=1.25in,clip,keepaspectratio]{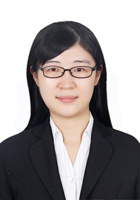}}]{Li Yu}
received the B.S. degree from Soochow University, Suzhou, China, in 2012, and the Ph.D. degree in electrical engineering and electronics from the University of Liverpool, Liverpool, U.K., in 2017. From 2017 to 2018, she was a Postdoctoral Researcher with the Department of Signal Processing, Tampere University of Technology, Tampere, Finland. Since 2018, she has been a Faculty Member with Nanjing University of Information Science and Technology, Nanjing, China. Her research interests include 3D computer vision, image and video processing and deep learning.
\end{IEEEbiography}

% \vspace{5pt}

\begin{IEEEbiography}[{\includegraphics[width=1in,height=1.25in,clip,keepaspectratio]{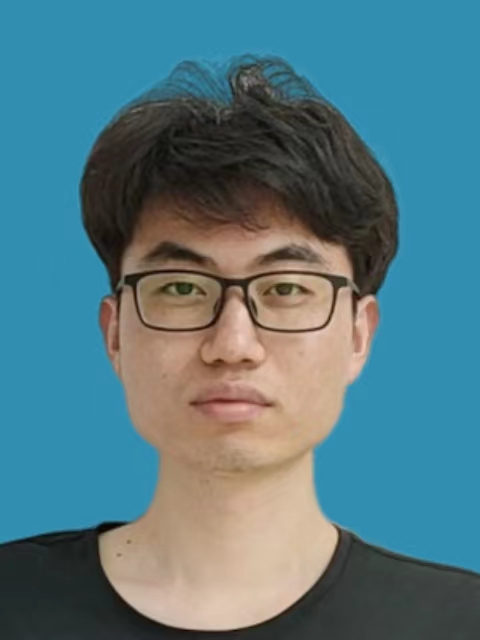}}]{Hongchao Zhong}
received the B.S. degree in Information security from the Nanjing University of Information Science and Technology (NUIST), China, in 2022, where he is currently pursuing the master's degree. His research interests include 3D computer vision and domain adaptation.
\end{IEEEbiography}

% \vspace{5pt}

\begin{IEEEbiography}[{\includegraphics[width=1in,height=1.25in,clip,keepaspectratio]{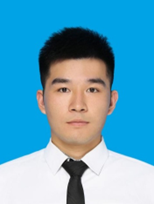}}]{Longkun Zou}
received the B.S. degree in software engineering from the School of Software Engineering, South China University of Technology, China, in 2016. He is currently pursuing the Ph.D. degree with the School of Electronic and Information Engineering, South China University of Technology, Guangzhou, China. His research interests include 3D computer vision and domain adaptation.
\end{IEEEbiography}

% \vspace{5pt}

\begin{IEEEbiography}[{\includegraphics[width=1in,height=1.25in,clip,keepaspectratio]{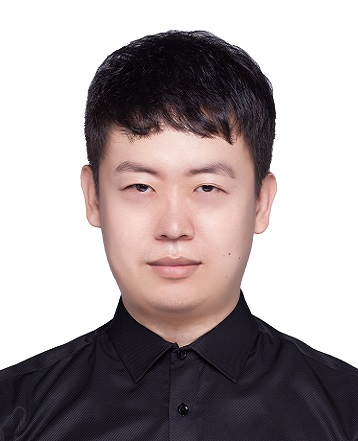}}]{Ke Chen} is currently an Associate Research Fellow with the the Peng Cheng Laboratory (PCL). Before joining PCL, he was an associate professor at the School of Electronic and Information Engineering, South China University of Technology, China; 
and a Postdoctoral Research Fellow with the Department of Signal Processing, Tampere University of Technology, Finland.
He received the B.E. degree in automation and the M.E. degree in software engineering from Sun Yat-sen University in 2007 and 2009, respectively, and the Ph.D. degree in computer vision from Queen Mary University of London in 2013. 
His research interests include computer vision, pattern recognition, neural dynamic modeling, and robotic inverse kinematics.
\end{IEEEbiography}

% \vspace{5pt}

\begin{IEEEbiography}[{\includegraphics[width=1in,height=1.25in,clip,keepaspectratio]{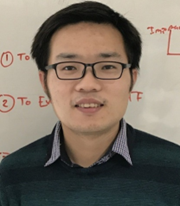}}]{Pan Gao} received  the Ph.D. degree in electronic engineering from University of Southern Queensland (USQ), Toowoomba, Australia, in 2017. 
Since 2016, he has been with the College of Computer Science and Technology, Nanjing University of Aeronautics and Astronautics, Nanjing, China, where he is currently an Associate Professor. From 2018 to 2019, he was a Postdoctoral Research Fellow at the School of Computer Science and Statistics, Trinity College Dublin, Dublin, Ireland, working on the V-SENSE project. He has authored or coauthored more than 60 publications in scientific journals and international conferences. His research interests include deep learning, computer vision, and artificial intelligence.

\end{IEEEbiography}

\vfill

\end{document}